\documentclass{article}

\usepackage{spconf,amsmath,graphicx}
\usepackage{amssymb} 
\usepackage{amsfonts} 
\usepackage{graphicx} 

\usepackage{multirow}
\usepackage{graphicx}
\usepackage{algorithm}
\usepackage{algorithmic}

\date{}

\usepackage{enumitem}

\usepackage{url}

\usepackage{hyperref}

\title{Are Objective Explanatory Evaluation metrics Trustworthy? An Adversarial Analysis}

\name{Prithwijit Chowdhury$^{\star}$ \qquad Mohit Prabhushankar$^{\star}$ \qquad Ghassan AlRegib$^{\star}$ \qquad Mohamed Deriche$^{\dagger}$}
\address{OLIVES at the Center for Signal and Information Processing\\School of Electrical and Computer Engineering\\
$^{\star}$Georgia Institute of Technology, Atlanta, GA 30332-0250\\
\{pchowdhury6, mohit.p, alregib\}@gatech.edu\\
$^{\dagger}$Ajman University, United Arab Emirates\\
\ m.deriche@ajman.ac.ae}

\date{}

\usepackage{enumitem}

\usepackage{url}

\usepackage{hyperref} 

\begin{document}

\onecolumn 

\begin{description}[labelindent=-1cm,leftmargin=1cm,style=multiline]


\item[\textbf{Citation}]{P. Chowdhury, M. Prabhushankar, G. AlRegib and M. Deriche, "Are Objective Explanatory Evaluation metrics Trustworthy? An Adversarial Analysis," in \textit{IEEE International Conference on Image Processing (ICIP)}, Abu Dhabi, UAE, Oct. 27 - 30, 2024.}

\item[\textbf{Review}]{Date of submission: Feb. 10, 2024\\Date of acceptance: Jun. 06, 2024}

\item[\textbf{Code}]{\url{https://github.com/olivesgatech/SHAPE}}



\item[\textbf{Copyright}]{\textcopyright 2024 IEEE. Personal use of this material is permitted. Permission from IEEE must be obtained for all other uses, in any current or future media, including reprinting/republishing this material for advertising or promotional purposes, creating new collective works, for resale or redistribution to servers or lists, or reuse of any copyrighted component of this work in other work. }


\item[\textbf{Keywords}]{Explainable AI, Causal Metrics, Adversarial Attacks, Visual Causality, Importance Maps} 

\item[\textbf{Contact}]{\href{mailto:pchowdhury6@gatech.edu}{pchowdhury6@gatech.edu}  OR 
 \href{mailto:mohit.p@gatech.edu}{mohit.p@gatech.edu} OR \href{mailto:alregib@gatech.edu}{alregib@gatech.edu} \\ \url{https://ghassanalregib.info/} \\ }
\end{description}

\thispagestyle{empty}
\newpage
\clearpage
\setcounter{page}{1}

\twocolumn


%

\maketitle
\begin{abstract}
Explainable AI (XAI) has revolutionized the field of deep learning by empowering users to have more trust in neural network models. The field of XAI allows users to probe the inner workings of these algorithms to elucidate their decision-making processes. The rise in popularity of XAI has led to the advent of different strategies to produce explanations, all of which only occasionally agree. Thus several objective evaluation metrics have been devised to decide which of these modules give the best explanation for specific scenarios. The goal of the paper is twofold: (i) we employ the notions of necessity and sufficiency from causal literature to come up with a novel explanatory technique called SHifted Adversaries using Pixel Elimination(SHAPE) which satisfies all the theoretical and mathematical criteria of being a valid explanation, (ii) we show that SHAPE is, infact, an adversarial explanation that fools causal metrics that are employed to measure the robustness and reliability of popular importance based visual XAI methods. Our analysis shows that SHAPE outperforms popular explanatory techniques like GradCAM and GradCAM++ in these tests and is comparable to RISE, raising questions about the sanity of these metrics and the need for human involvement for an overall better evaluation.

\end{abstract}

\section{Introduction}
\label{sec:intro}
\begin{figure}[h]
  \centering
  \includegraphics[width=\columnwidth]{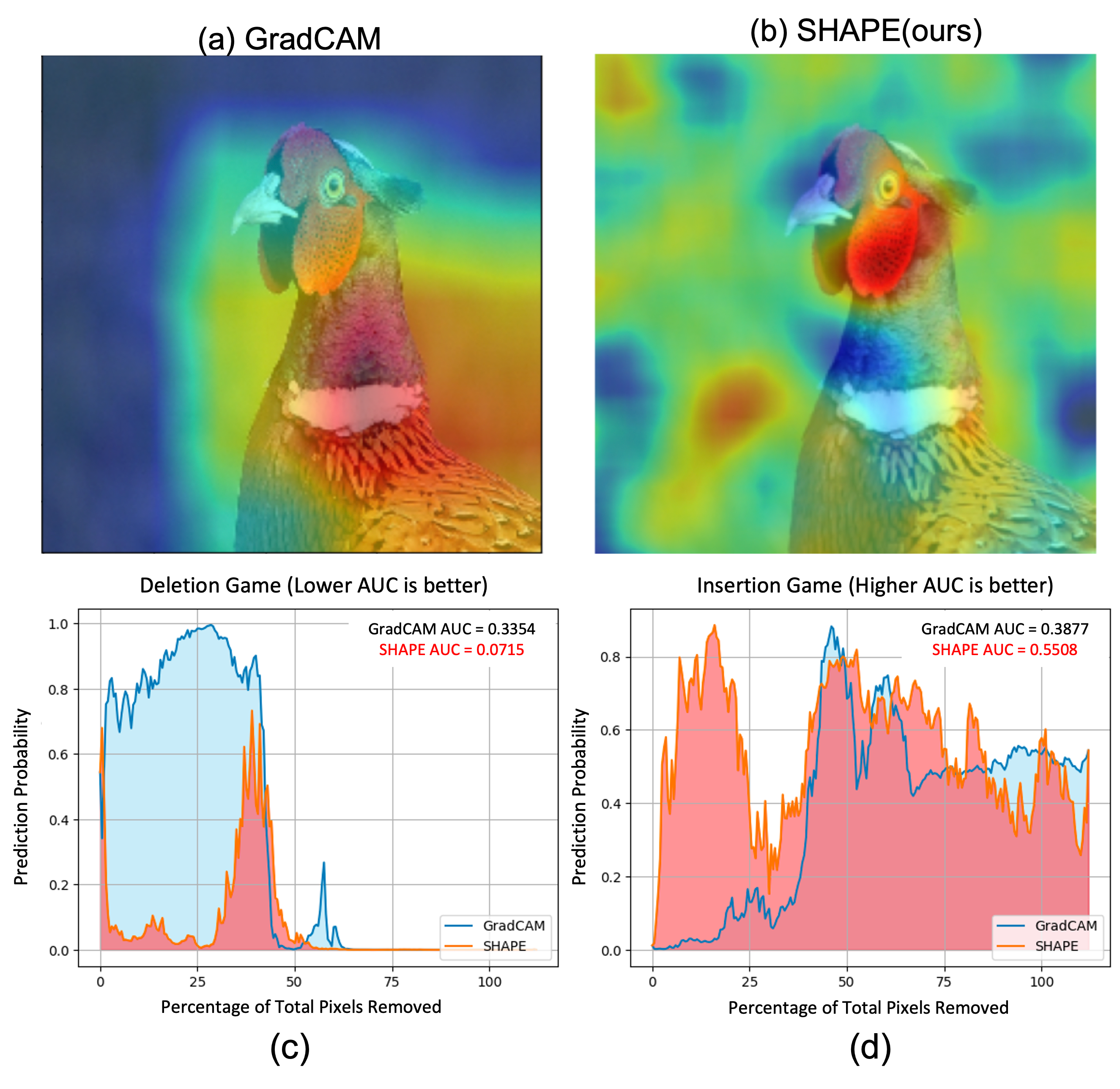} \vspace{-0.8cm} 
  \caption{Even though the adversarial explanation generated by SHAPE (b) is much less comprehensible as compared to the GradCAM explanation (a), it ourperforms the later in both the causal metric test (c) \& (d) by a significant margin thus claiming to be the more robust and reliable among the two.}
  \label{fig:front}
\end{figure}
The recent success of neural networks (NN) has resulted in a notable expansion in artificial intelligence (AI) research. Despite it’s popularity, there is still much to learn about how a given neural network makes decisions, and when and if it can be trusted. It's crucial that decision-making models are transparent in fields like medical diagnosis \cite{prabhushankar2022olives}, autonomous driving \cite{kokilepersaud2023focal}, geophysics \cite{chowdhury2023counterfactual} and others where decisions can have dire repercussions. The field of interpretable machine learning (or Explainable Artificial Intelligence, or XAI) focuses on post-hoc, model-independent explanation tools to overcome this problem \cite{alregib2022explanatory, prabhushankar2022introspective, prabhushankar2020contrastive}.

An intuitive approach to generate explanations for a NN is to visualize a certain quantity of interest, e.g. importance of input features or learned weights. In case of Convolution Neural Networks (CNNs) which deal with image data, this can be represented in the form of explanation maps, highlighting regions of pixels in a certain order of importance. Two common methods to generate these explanatory importance maps are: gradient based (e.g. GradCAM \cite{selvaraju2017grad}, GradCAM++\cite{chattopadhay2018grad}, CausalCAM \cite{prabhushankar2021extracting}, ContrastCAM \cite{DBLP:journals/corr/abs-2008-00178} etc.)  and perturbation based (e.g. RISE\cite{petsiuk2018rise} and LIME \cite{ribeiro2016should}). Despite the potential of these XAI strategies, the discrepancies and the disagreement among them raise concerns about their reliability \cite{murdoch2019definitions, krishna2022disagreement, prabhushankar2024voice}. Hence, scholars have come up with metrics and benchmarks to test the robustness of these XAI strategies themselves. Recent work \cite{kommiya2021towards, ramon2020comparison} has argued that a convincing importance-ranking explanation should always satisfy the desirable properties of causality and answer the question: \textit{How necessary and sufficient are the important regions?} \cite{petsiuk2018rise} has formulated this question using their two tests: insertion game and deletion game. As the names suggest, they introduce and remove regions of an image according to their importance proposed by the XAI strategy to see how fast or slow the model changes the prediction probability of the concerned class. However since most explanation strategies tend to approximate the idea of correlation as causation in their algorithms, they fail when subjected to the tests of these causal metrics. This raises a follow-up question in this ongoing conversation: \textit{Are causal metrics the desirable test to assess the robustness of XAI modules?}

In our work we take the concept of adversarial attack \cite{akhtar2018threat} and extend them to explanations. Adversarial attacks are originally designed to deceive or mislead machine learning models by introducing carefully crafted perturbations in the input data, thereby questioning the reliability of deep learning models. In this paper, we generate adversarial explanations to test the sanity of the causal XAI evaluation metrics. Fig.\ref{fig:front}. shows the explanatory maps as well as the insertion and deletion performance of explanations generated by GradCAM (Fig.\ref{fig:front}a.) and our adversarial method (SHAPE) (Fig.\ref{fig:front}b.) for the same class prediction (Rooster). Subjectively, GradCAM is a better explanation for why the network predicted a Rooster. Objectively, GradCAM's explanation should be considered \emph{superior} than SHAPE's if it's Area Under Curve (AUC) for the insertion test is higher. Similarly, AUC must be lower in the deletion test to be considered superior. In both cases (Fig.\ref{fig:front}c.\& d.), however, our SHAPE explanatory technique outperforms GradCAM, contrarily to our subjective expectation. Hence, Fig.~\ref{fig:front} reveals the flaws in objective evaluation methods and raises concerns about how explanations are produced in general. In summary the contributions of this paper include:

\begin{enumerate}
    \item Generating model-faithful adversarial explanation generations based on the idea of necessity and sufficiency.
    \item Evaluating the validity of the current objective causal metrics used to test the robustness and reliability of importance map based XAI modules.
\end{enumerate}

\section{Background}
\label{sec:back}

\subsection{Causal Definitions of Importance: Necessity and Sufficiency}

Necessity and sufficiency are concepts that have been extensively explored in philosophy, encompassing various logical, probabilistic, and causal interpretations. (\cite{swartz1997concepts}) defines necessity and sufficiency as follows:

\noindent\textbf{\textit{Necessary condition:}} A condition \textbf{A} is said to be necessary for a condition \textbf{B}, iff the falsity of \textbf{A} guarantees the falsity of \textbf{B}.
\textit{\textbf{Example:}``Air is necessary for human life."} 

\noindent\textbf{\textit{Sufficient condition:}} A condition \textbf{A} is said to be sufficient for a condition \textbf{B}, iff the truth  of \textbf{A} guarantees the truth of \textbf{B}. 
\textit{\textbf{Example:}``Being a mammal is sufficient to have a spine."}

The author in~\cite{pearl2009causality} has hypothesised that these two concepts from causal analysis help to assess the degree of importance or significance displayed by any cause of an event. Recent work \cite{kommiya2021towards, watson2021local, galhotra2021explaining} has claimed that a models prediction and its relation to its input space can be understood using the same logic of cause and effect. Intervention is done by removing or introducing features from the input space and studying how the model's behavior changes from the non-intervened case \cite{chowdhury2023explaining}. A necessary feature will change a models decision when it is removed and a sufficient one will do so when it is introduced. by formulating these two causal concepts using probabilistic approaches we can attempt to calculate and quantify the importance of input features for a specific models prediction. Attribute specific XAI modules are designed to do just that.

\subsection{Visual Explanations: Importance Maps}

Importance maps are a type of visual XAI tool used in deep learning and computer vision to determine which regions of an input image have the greatest influence on a neural network model's decision \cite{ullah2020brief}.

Methods to generate importance maps can be broadly classified into two techniques:

\begin{enumerate}
    \item \textbf{Gradient-based Backpropagation Method}: Gradient-based backpropagation techniques for importance maps in computer vision and image processing entail iteratively adjusting a neural network's weights to minimize a predetermined loss function that measures the discrepancy between predicted importance maps and ground truth annotations. Some examples of such methods are GradCAM \cite{selvaraju2017grad} GradCAM++ \cite{chattopadhay2018grad}  CausalCAM \cite{prabhushankar2021extracting}, ContrastCAM \cite{DBLP:journals/corr/abs-2008-00178} and Guided Backpropagation \cite{nie2018theoretical, lee2023probing}.

    \item \textbf{Perturbation-based Forward Propagation  Method}: These techniques entail systematically modifying the input images and recording how these modifications impact the neural network model's output \cite{ivanovs2021perturbation}. The method RISE provides a way to create explanatory maps that emphasize the important of the input data for the model's decision-making process by masking different regions of an image and examining how the remaining area affects the model's predictions. 
\end{enumerate}

\subsection{Causal Metrics for Evaluation of Explanations}
\label{metrics}
With the emergence of numerous Explainable Artificial Intelligence (XAI) methods focused on importance maps, there's an increased demand for objective evaluation strategies that determine which properties yield accurate and reliable explanations. Motivated by the work of \cite{fong2017interpretable}, the authors in \cite{petsiuk2018rise} formulate the definitions of necessity and sufficiency to establish two causal metrics for explanation evaluation: deletion and insertion.
\begin{enumerate}
    \item \textbf{Deletion:} The deletion metric quantifies the definition of necessity which is based on the logical assumption that eliminating the "cause" will compel the base model to change its decision. In particular, when increasing numbers of significant pixels are eliminated—the significance being derived from the explanation map—this metric quantifies a decline in the likelihood of the anticipated class. A good explanation is indicated by a steep drop and consequently a low area under the probability curve or AUC (as a function of the fraction of eliminated pixels). Based on this conclusion, SHAPE in Fig.\ref{fig:front} is a better strategy for class explaining the class rooster because it has a smaller AUC (a 76.66\% decrease) than GradCAM.
    \item \textbf{Insertion:} In contrast, the insertion metric employs a complimentary methodology in order to calculate the sufficiency of the explanation. As more and more pixels are added, it calculates the likelihood increase, with a broader curve and higher AUC suggesting a more compelling explanation. The insertion AUC of SHAPE for Fig.\ref{fig:front} is almost 50\% greater than GradCAM establishing it as a better explanation according to both the causal metrics.
    
\end{enumerate}

\subsection{Adversarial Explanations: A New Robustness Test}

In deep learning, especially computer vision, an adversarial attack is defined as the purposeful manipulation of the input image in order to trick the machine learning model to produce false decisions \cite{akhtar2018threat, tabacof2016exploring}. These adversarial images have barely any imperceptible changes in them while making wrong predictions thereby questioning the softmax probability metric. We have adopted a converse approach to create our adversarial explanations. Objectively, our explanations employ the idea of necessary features to produce the importance maps and have better insertion and deletion game scores than other explanatory maps present in current literature. However subjectively, these explanations are not human interpretable and thus don't contribute to increase trust between the user and the neural network- a condition any dependable XAI module needs to satisfy.  Hence these adversarial explanations take a reverse approach to raise questions about the robustness of these causal metrics being suitable evaluation tests for importance map based XAI approaches.

\section{Methodology}
\label{sec:method}

In this work we propose to leverage the two notions of causality, necessity and sufficiency, to come up with a framework to generate adversarial explanations. Our method is simultaneously faithful to the model itself and mathematically sound when generating visual explanations. However, our explanations are adversarial - they are almost entirely incomprehensible by humans (Fig. \ref{fig:front} \& \ref{fig:ADV_EX}) thus voiding the whole idea of an explanation helping its user understand the inner workings of the model while increasing trust in its decision-making process.  

Our algorithm is an inverse approach to the popular perturbation based method, RISE \cite{petsiuk2018rise}. RISE explanations quantify the sufficiency of a region in the image by masking out all other regions and allowing the model to make predictions on the area of interest only. Our method creates explanations by quantifying the shift in model response when a specific portion of pixels are eliminated. In other words, our explanations quantify the necessity of a region of pixels.

\subsection{SHifted Adversaries using Pixel Elimination (SHAPE)}

\begin{figure*}[t!] 
  \centering
  \includegraphics[width=\textwidth]{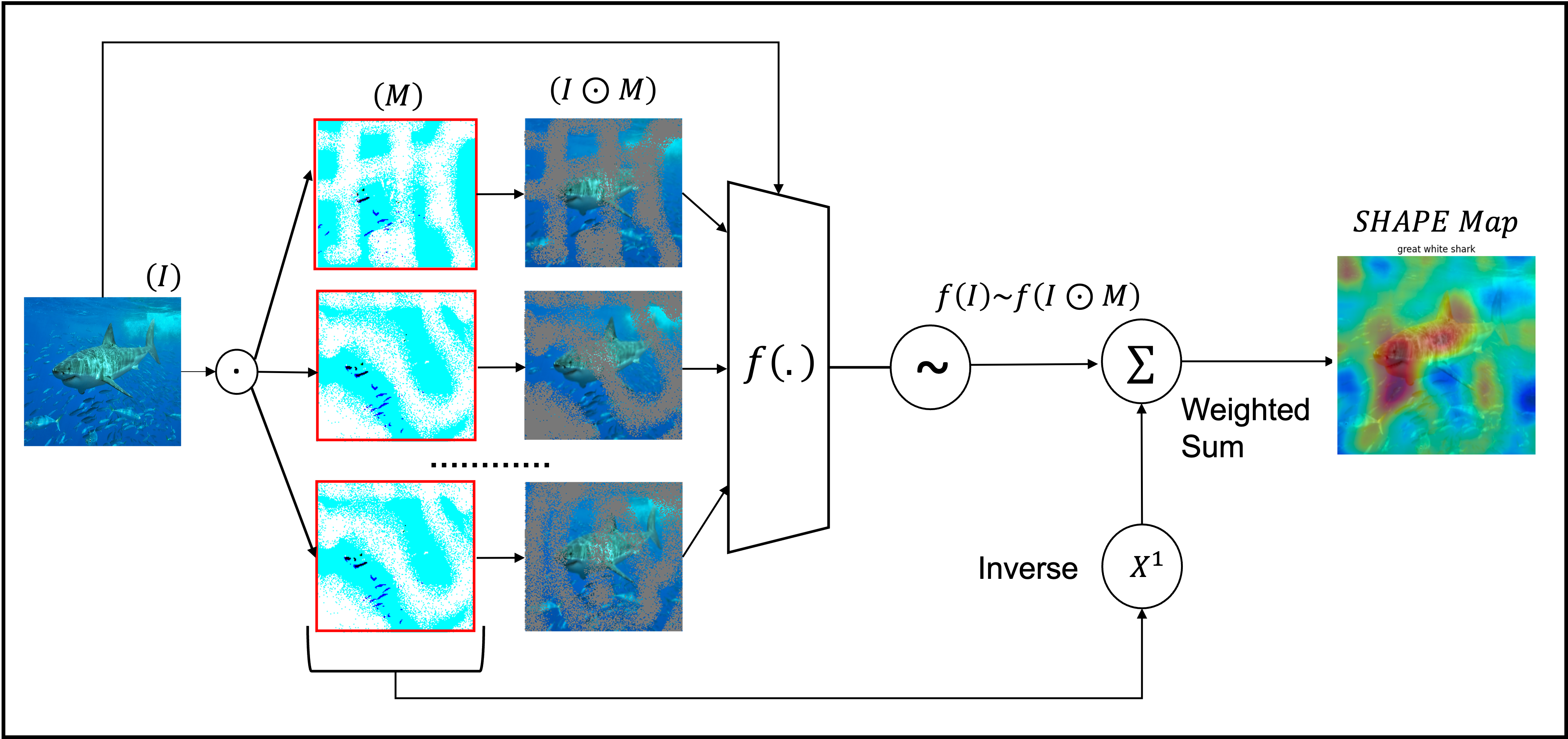} 
  \caption{Overview of \textbf{SHAPE} to generate adversarial explanations: Input image $I$ is element-wise multiplied by the random masks $M_i$ and are fed into the model $f$ along with the original image to calculate the change in prediction scores.The importance map is a weighted sum of masks where the weights for each mask is its corresponding change in probability scores. }
  \label{fig:method}
\end{figure*}

Based on the random masking method introduced in \cite{petsiuk2018rise}, we use the definition of necessary cause to define the importance of pixels $\lambda$ as the expected change in prediction probability from the original image $(I)$ over all masks $M$ on the condition that the pixels are not observed, \textit{i.e.}, $M(\lambda) = 0$:

\begin{equation}\label{eq:eq1}
N_{I, f}(\lambda)=\mathbb{E}_M[f(I) \sim f(I \odot M) \mid M(\lambda)=0]
\end{equation}
where $I \odot M$ refers to the image with the mask applied.

For a summation over all masks $m$, Eq. \ref{eq:eq1} is:
\begin{align}\label{eq:eq2}
N_{I, f}(\lambda) \\
&= \sum_m [f(I) \sim f(I \odot m) ] P[M=m \mid M(\lambda)=0] \nonumber
\end{align}
which can be written as:
\begin{align}\label{eq:eq3}
\frac{1}{P[M(\lambda)=0]} \sum_m [f(I) \sim f(I \odot m) ] P[M=m, M(\lambda)=0]
\end{align}
Here,
\begin{align}\label{eq:eq4}
P[M=m, M(\lambda)=0] &= 
\begin{cases}
P[M=m], & \text{if } m(\lambda)=0 \\
0, & \text{if } m(\lambda)=1
\end{cases} \nonumber \\
&= (1 - m(\lambda)) P[M=m]
\end{align}

Substituting $P[M=m, M(\lambda)=0]$ from (\ref{eq:eq4}) in (\ref{eq:eq3}) we get:
{\scriptsize
\begin{equation}\label{eq:eq5}
N_{I, f}(\lambda) = \frac{1}{P[M(\lambda)=0]} \sum_m [f(I) \sim f(I \odot m) ] \cdot (1-m(\lambda)) \cdot P[M=m]
\end{equation}
}

We then use Monte Carlo's approximation to reduce the summation to a weighted average of the mass samples $M(\lambda)$:

\begin{equation}\label{eq:eq6}
\small
N_{I, f}(\lambda) \stackrel{\mathrm{MC}}{\approx} \frac{1}{P[M(\lambda)=0] \cdot N} \sum_{i=1}^N [f(I) \sim f(I \odot M_i) ] \cdot(1- M_i(\lambda)) .
\end{equation}

The final $N_{I, f}(\lambda)$ is an approximation of region-wise necessity score for a particular class prediction for an image $I$ and using model $f$ (See Fig \ref{fig:method}).


\section{Experiments and Observations}
\label{sec:exp}

\subsection{Experimental Setup for Adversarial Explanation Generation}
Similar to \cite{petsiuk2018rise} we generate the masks for pixel deletion. Smaller binary masks are randomly upsampled using bilinear interpolation to generate masks of larger resolution to ensure smooth importance map generation. The dimensions of binary masks have been kept at a constant size of $7X7$ pixels with pixel restoration ($\lambda = 1$) probability being $0.1$. Using this binary mask we generate $6000$ large masks to be used in all our experiments. We generate adversarial explanations for images inferred using a ResNet50, RestNet101, and VGG16 model, all pretrained on the ImageNet dataset. 
\begin{figure}
  \centering
  \includegraphics[width=\columnwidth]{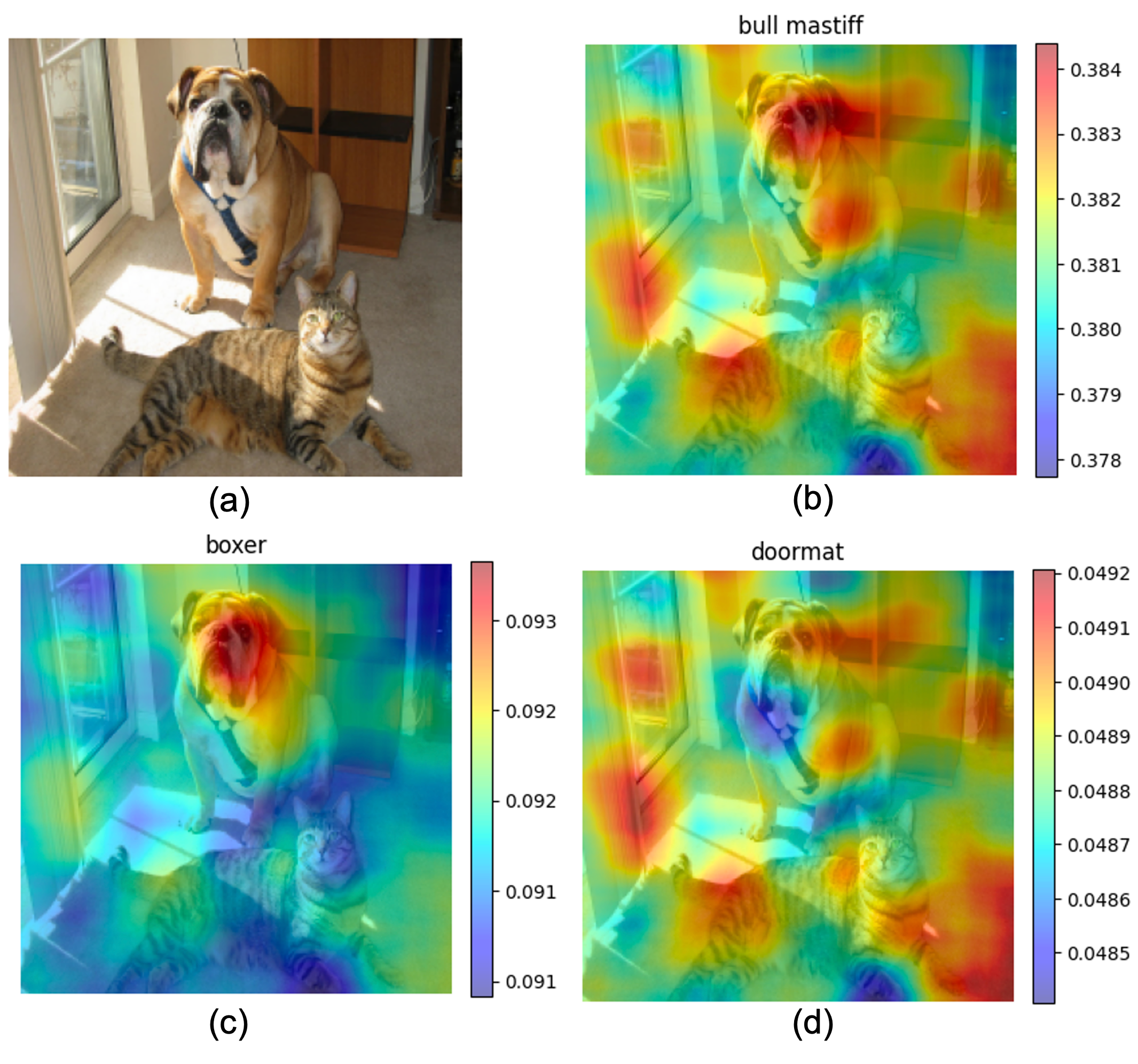} 
  \caption{\textbf{SHAPE} Maps for image (a) for a RestNet101 model. (b) shows the importance map for class Bull mastiff (prediction accuracy = $38.42\%$), (c) shows the importance map for class Tiger cat (prediction accuracy = $9.41\%$) and (d) shows the map for class Tabby (prediction accuracy = $04.97\%$).}
  \label{fig:ADV_EX}
\end{figure}

\begin{figure*}[t]
  \centering
  \includegraphics[width=\textwidth]{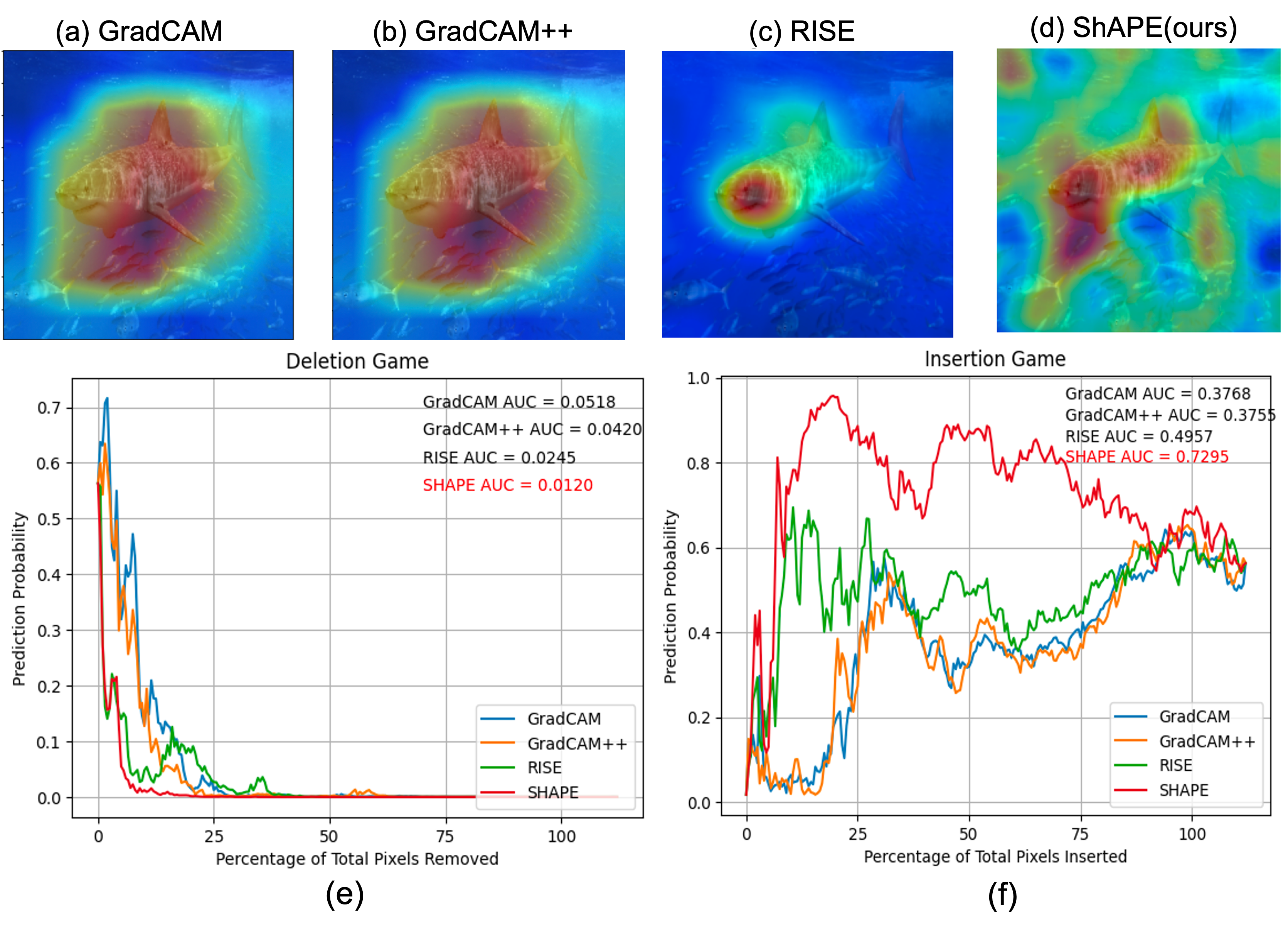} 
  \caption{CAM for prediction ``great white shark" using (a) GradCAM, (b) GradCAM++, (c) RISE and d) SHAPE(ours). SHAPE significantly outperforms all other methods in both (e) deletion (\textit{lower AUC is better}) and (f) deletion (\textit{higher AUC is better}) game}
  \label{fig:comp}
\end{figure*}

Fig.\ref{fig:ADV_EX} shows some of the adversarial explanations generated using \textbf{SHAPE} scores for different class predictions. Subjectively, only the SHAPE map for boxer in (c) seems to be the most human understandable explanation, while the one for bull mastiff as well as doormat are almost indistinguishable and simply represented by splotches of importance region spread throughout the image.

Subjectively, most of the visual explanations generated using \textbf{SHAPE} are not human interpretable (See Fig \ref{fig:comp} for another example). However they are sound on the basis of the fact that this map shows what region a model considers important or necessary during generation of it's prediction for a certain class. Thus this method satisfies the definition of an activation based visual explanation strategy: \textit {it should provide insights into which regions of an input image are important for a neural network's decision-making process.} Hence for all experiments and mathematical analysis \textbf{SHAPE} can be used as a valid explanation.

\subsection{Faithfulness Test of Causal Evaluation Metrics }

We notice \textbf{SHAPE}, our adversarial methods performs better on both the deletion and insertion game for the particular example in Fig.\ref{fig:comp} than all the other three explanation methods (GradCAM, GradCAM++ and RISE). It shows almost a 50\% better AUC improvement in both metrics when compared to the second best explanation strategy, RISE. Thus according to the causal metric evaluation \textbf{SHAPE} explantions are the best and most faithful to explain this model's decision making process. However human inspection of the explanation map proves otherwise. This holds true for several other test images from the ImageNet dataset as well, with severely misconstrued importance maps displaying the best performances in both the causal tests (also Fig.\ref{fig:front}). 

\begin{table}[]
\centering
\resizebox{\columnwidth}{!}{%
\begin{tabular}{|c|cc|cc|cc|}
\hline
\multirow{2}{*}{Method} & \multicolumn{2}{c|}{ResNet50}      & \multicolumn{2}{c|}{ResNet101}      & \multicolumn{2}{c|}{VGG16}         \\ \cline{2-7} 
 &
  \multicolumn{1}{c|}{Insertion} &
  Deletion &
  \multicolumn{1}{c|}{Insertion} &
  Deletion &
  \multicolumn{1}{c|}{Insertion} &
  Deletion \\ \hline
GradCAM                 & \multicolumn{1}{c|}{0.684} & 0.223 & \multicolumn{1}{c|}{0.6871} & 0.174 & \multicolumn{1}{c|}{0.614} & 0.118 \\ \hline
GradCAM++               & \multicolumn{1}{c|}{0.712} & 0.209 & \multicolumn{1}{c|}{0.701}  & 0.166 & \multicolumn{1}{c|}{0.642} & 0.134 \\ \hline
RISE &
  \multicolumn{1}{c|}{0.769} &
  \textbf{0.091} &
  \multicolumn{1}{c|}{0.788} &
  0.150 &
  \multicolumn{1}{c|}{\textbf{0.755}} &
  \textbf{0.099} \\ \hline
ShAPE(Ours) &
  \multicolumn{1}{c|}{\textbf{0.771}} &
  0.104 &
  \multicolumn{1}{c|}{\textbf{0.844}} &
  \textbf{0.088} &
  \multicolumn{1}{c|}{0.731} &
  0.101 \\ \hline
\end{tabular}%
}
\caption{Mean insertion (higher is better) and deletion (lower is better) evaluation scores for GradCAM, GradCAM++, RISE and SHAP for models: ResNet50, RestNet101 and VGG16 for same number of test images randomly sampled from ImageNET DET test set.}
\label{tab:results}
\end{table}

Table \ref{tab:results} shows that for samples queried randomly from ImageNet DET test the mean deletion and insertion scores for \textbf{SHAPE} makes it a much more trustworthy explanation than GradCAM and GradCAM++ for all three (ResNet50, ResNet101 and VGG16) models and even comparable to RISE.

\
\subsection{Discussion}

Although \textbf{SHAPE} doesn't overload the explanation map with high importance scores like GradCAM or GradCAM++ does in Fig.\ref{fig:comp} a.\& b., it fools both the insertion and deletion metrics despite highlighting important pixel regions outside the area of interest (where the object is actually located) for the particular class prediction. This can be attributed to the fact that the \emph{important} regions shown by these XAI modules aren't always the same on a causal scale. This is because causal features cannot be extracted exclusively from pixels \cite{prabhushankar2021extracting}. Unlike in structured data where causal relationships might be more discernible, images present a complex web of interconnected pixels, making it difficult to isolate and manipulate individual causal factors. While neural networks excel at learning complex patterns and making predictions based on correlations within the data, they lack the inherent capacity to infer causality without explicit guidance. While predictive features may suffice for accurate predictions in normal circumstances, causal features are indispensable for maintaining accuracy even in the face of corruptions. Consequently even minor alterations can trigger substantial shifts in the representation. Hence, causal metrics like insertion and deletion can fail to hold as baselines for sanity checks of these explanations, which are but a pixel level approximation of what causal features might represent.

\section{Conclusion}
\label{sec:conc}
In this paper we propose a novel method to create adversarial explanations by formulating the definition of necessary features, as introduced by causality, and quantifying the changes in model prediction. We use this adversarial strategy to study the rationality behind using casual evaluation metrics like insertion and deletion of pixels as a valid technique for comparing the robustness of importance based visual explanation algorithms. Our adversarial technique outperforms existing visual explanations in these evaluation metrics. Hence, our work motivates the need for devising objective explanatory evaluation techniques that matches subjective human opinions. Until such metrics are devised, human intervention and assessment is necessary for proper evaluation of visual explanations.


\bibliographystyle{IEEEbib}
\bibliography{icip2024}

\begin{thebibliography}{10}

\bibitem{prabhushankar2022olives}
Mohit Prabhushankar, Kiran Kokilepersaud, Yash-yee Logan, Stephanie Trejo~Corona, Ghassan AlRegib, and Charles Wykoff,
\newblock ``Olives dataset: Ophthalmic labels for investigating visual eye semantics,''
\newblock {\em Advances in Neural Information Processing Systems}, vol. 35, pp. 9201--9216, 2022.

\bibitem{kokilepersaud2023focal}
Kiran Kokilepersaud, Yash-Yee Logan, Ryan Benkert, Chen Zhou, Mohit Prabhushankar, Ghassan AlRegib, Enrique Corona, Kunjan Singh, and Mostafa Parchami,
\newblock ``Focal: A cost-aware video dataset for active learning,''
\newblock in {\em 2023 IEEE International Conference on Big Data (BigData)}. IEEE, 2023, pp. 1269--1278.

\bibitem{chowdhury2023counterfactual}
Prithwijit Chowdhury, Ahmad Mustafa, Mohit Prabhushankar, and Ghassan AlRegib,
\newblock ``Counterfactual uncertainty for high dimensional tabular dataset,''
\newblock in {\em SEG International Exposition and Annual Meeting}. SEG, 2023, pp. SEG--2023.

\bibitem{alregib2022explanatory}
Ghassan AlRegib and Mohit Prabhushankar,
\newblock ``Explanatory paradigms in neural networks: Towards relevant and contextual explanations,''
\newblock {\em IEEE Signal Processing Magazine}, vol. 39, no. 4, pp. 59--72, 2022.

\bibitem{prabhushankar2022introspective}
Mohit Prabhushankar and Ghassan AlRegib,
\newblock ``Introspective learning: A two-stage approach for inference in neural networks,''
\newblock {\em Advances in Neural Information Processing Systems}, vol. 35, pp. 12126--12140, 2022.

\bibitem{prabhushankar2020contrastive}
Mohit Prabhushankar, Gukyeong Kwon, Dogancan Temel, and Ghassan AlRegib,
\newblock ``Contrastive explanations in neural networks,''
\newblock in {\em 2020 IEEE International Conference on Image Processing (ICIP)}. IEEE, 2020, pp. 3289--3293.

\bibitem{selvaraju2017grad}
Ramprasaath~R Selvaraju, Michael Cogswell, Abhishek Das, Ramakrishna Vedantam, Devi Parikh, and Dhruv Batra,
\newblock ``Grad-cam: Visual explanations from deep networks via gradient-based localization,''
\newblock in {\em Proceedings of the IEEE international conference on computer vision}, 2017, pp. 618--626.

\bibitem{chattopadhay2018grad}
Aditya Chattopadhay, Anirban Sarkar, Prantik Howlader, and Vineeth~N Balasubramanian,
\newblock ``Grad-cam++: Generalized gradient-based visual explanations for deep convolutional networks,''
\newblock in {\em 2018 IEEE winter conference on applications of computer vision (WACV)}. IEEE, 2018, pp. 839--847.

\bibitem{prabhushankar2021extracting}
Mohit Prabhushankar and Ghassan AlRegib,
\newblock ``Extracting causal visual features for limited label classification,''
\newblock in {\em 2021 IEEE International Conference on Image Processing (ICIP)}. IEEE, 2021, pp. 3697--3701.

\bibitem{DBLP:journals/corr/abs-2008-00178}
Mohit Prabhushankar, Gukyeong Kwon, Dogancan Temel, and Ghassan AlRegib,
\newblock ``Contrastive explanations in neural networks,''
\newblock {\em CoRR}, vol. abs/2008.00178, 2020.

\bibitem{petsiuk2018rise}
Vitali Petsiuk, Abir Das, and Kate Saenko,
\newblock ``Rise: Randomized input sampling for explanation of black-box models,''
\newblock {\em arXiv preprint arXiv:1806.07421}, 2018.

\bibitem{ribeiro2016should}
Marco~Tulio Ribeiro, Sameer Singh, and Carlos Guestrin,
\newblock ``" why should i trust you?" explaining the predictions of any classifier,''
\newblock in {\em Proceedings of the 22nd ACM SIGKDD international conference on knowledge discovery and data mining}, 2016, pp. 1135--1144.

\bibitem{murdoch2019definitions}
W~James Murdoch, Chandan Singh, Karl Kumbier, Reza Abbasi-Asl, and Bin Yu,
\newblock ``Definitions, methods, and applications in interpretable machine learning,''
\newblock {\em Proceedings of the National Academy of Sciences}, vol. 116, no. 44, pp. 22071--22080, 2019.

\bibitem{krishna2022disagreement}
Satyapriya Krishna, Tessa Han, Alex Gu, Javin Pombra, Shahin Jabbari, Steven Wu, and Himabindu Lakkaraju,
\newblock ``The disagreement problem in explainable machine learning: A practitioner's perspective,''
\newblock {\em arXiv preprint arXiv:2202.01602}, 2022.

\bibitem{prabhushankar2024voice}
Mohit Prabhushankar and Ghassan AlRegib,
\newblock ``Voice: Variance of induced contrastive explanations to quantify uncertainty in neural network interpretability,''
\newblock {\em arXiv preprint arXiv:2406.00573}, 2024.

\bibitem{kommiya2021towards}
Ramaravind Kommiya~Mothilal, Divyat Mahajan, Chenhao Tan, and Amit Sharma,
\newblock ``Towards unifying feature attribution and counterfactual explanations: Different means to the same end,''
\newblock in {\em Proceedings of the 2021 AAAI/ACM Conference on AI, Ethics, and Society}, 2021, pp. 652--663.

\bibitem{ramon2020comparison}
Yanou Ramon, David Martens, Foster Provost, and Theodoros Evgeniou,
\newblock ``A comparison of instance-level counterfactual explanation algorithms for behavioral and textual data: Sedc, lime-c and shap-c,''
\newblock {\em Advances in Data Analysis and Classification}, vol. 14, pp. 801--819, 2020.

\bibitem{akhtar2018threat}
Naveed Akhtar and Ajmal Mian,
\newblock ``Threat of adversarial attacks on deep learning in computer vision: A survey,''
\newblock {\em Ieee Access}, vol. 6, pp. 14410--14430, 2018.

\bibitem{swartz1997concepts}
Norman Swartz,
\newblock ``The concepts of necessary conditions and sufficient conditions,''
\newblock {\em Department of Philosophy Simon Fraser University}, 1997.

\bibitem{pearl2009causality}
Judea Pearl,
\newblock {\em Causality},
\newblock Cambridge university press, 2009.

\bibitem{watson2021local}
David~S Watson, Limor Gultchin, Ankur Taly, and Luciano Floridi,
\newblock ``Local explanations via necessity and sufficiency: Unifying theory and practice,''
\newblock in {\em Uncertainty in Artificial Intelligence}. PMLR, 2021, pp. 1382--1392.

\bibitem{galhotra2021explaining}
Sainyam Galhotra, Romila Pradhan, and Babak Salimi,
\newblock ``Explaining black-box algorithms using probabilistic contrastive counterfactuals,''
\newblock in {\em Proceedings of the 2021 International Conference on Management of Data}, 2021, pp. 577--590.

\bibitem{chowdhury2023explaining}
Prithwijit Chowdhury, Mohit Prabhushankar, and Ghassan AlRegib,
\newblock ``Explaining explainers: Necessity and sufficiency in tabular data,''
\newblock in {\em NeurIPS 2023 Second Table Representation Learning Workshop}, 2023.

\bibitem{ullah2020brief}
Inam Ullah, Muwei Jian, Sumaira Hussain, Jie Guo, Hui Yu, Xing Wang, and Yilong Yin,
\newblock ``A brief survey of visual saliency detection,''
\newblock {\em Multimedia Tools and Applications}, vol. 79, pp. 34605--34645, 2020.

\bibitem{nie2018theoretical}
Weili Nie, Yang Zhang, and Ankit Patel,
\newblock ``A theoretical explanation for perplexing behaviors of backpropagation-based visualizations,''
\newblock in {\em International conference on machine learning}. PMLR, 2018, pp. 3809--3818.

\bibitem{lee2023probing}
Jinsol Lee, Charlie Lehman, Mohit Prabhushankar, and Ghassan AlRegib,
\newblock ``Probing the purview of neural networks via gradient analysis,''
\newblock {\em IEEE Access}, vol. 11, pp. 32716--32732, 2023.

\bibitem{ivanovs2021perturbation}
Maksims Ivanovs, Roberts Kadikis, and Kaspars Ozols,
\newblock ``Perturbation-based methods for explaining deep neural networks: A survey,''
\newblock {\em Pattern Recognition Letters}, vol. 150, pp. 228--234, 2021.

\bibitem{fong2017interpretable}
Ruth~C Fong and Andrea Vedaldi,
\newblock ``Interpretable explanations of black boxes by meaningful perturbation,''
\newblock in {\em Proceedings of the IEEE international conference on computer vision}, 2017, pp. 3429--3437.

\bibitem{tabacof2016exploring}
Pedro Tabacof and Eduardo Valle,
\newblock ``Exploring the space of adversarial images,''
\newblock in {\em 2016 international joint conference on neural networks (IJCNN)}. IEEE, 2016, pp. 426--433.

\end{thebibliography}

\end{document}